\documentclass[runningheads]{llncs}
\usepackage{graphicx}
\usepackage{comment}
\usepackage{amsmath,amssymb} 
\usepackage{color}

\usepackage{url} 

\begin{document}

\pagestyle{headings}
\mainmatter
\title{Abiotic Stress Prediction from RGB-T Images of Banana Plantlets} 

\titlerunning{Abiotic Stress Prediction from RGB-T Images of Banana Plantlets}

\author{Sagi Levanon\inst{1} \and
Oshry Markovich\inst{2} \and
Itamar Gozlan\inst{1} \and
Ortal Bakhshian\inst{2} \and
Alon Zvirin\inst{1} \and
Yaron Honen\inst{1} \and
Ron Kimmel\inst{1}}
\authorrunning{S. Levanon et al.}
%
\institute{Computer Science Department, Technion - Israel Institute of Technology
\url{http://www.cs.technion.ac.il}
 \and
Rahan Meristem (1998) Ltd.
\url{http://www.rahan.co.il}
}

\maketitle

\begin{abstract}
Prediction of stress conditions is important for monitoring plant growth stages, disease detection, and assessment of crop yields. 
Multi-modal data, acquired from a variety of sensors, offers diverse perspectives and is expected to benefit the prediction process.
We present several methods and strategies for abiotic stress prediction in banana plantlets, on a dataset acquired during a two and a half weeks period, of plantlets subject to four separate water and fertilizer treatments.
The dataset consists of RGB and thermal images, taken once daily of each plant. 
Results are encouraging, in the sense that neural networks exhibit high prediction rates (over $90\%$ amongst four classes), in cases where there are hardly any noticeable features distinguishing the treatments, much higher than field experts can supply.  

\end{abstract}

\section{Introduction}
Stress conditions in plants are usually divided into two categories - biotic, induced by biological factors such as fungi or bacteria, and abiotic, caused by climatic conditions such as heat or drought.
In this paper we deal with prediction of abiotic stress, by designing an experiment for creating and analysing four separate treatments of water and fertilizer conditions for banana plantlets in a greenhouse. 
The specific species is known as {\em Musa acuminata}, considered one of the earliest domesticated plants \cite{denham2003origins,perrier2011multidisciplinary,lebot1999biomolecular,de2009bananas,li2013origins}, cultivated and hybridized extensively in recent centuries for mass production of human edible clones \cite{sheperd1999cytogenetics,khayat2011genetics,fuller2009banana}.
Banana is the fourth largest fruit crop in the world, considered a staple food in many countries \cite{surendar2013water}.
However, one of the major limitations to it's productivity is water stress. 
Sensitivity to reduction in soil moisture is reflected in dwindled growth due to reduced stomatal conductance and leaf size which eventually may lead to plant death \cite{kallarackal1990water}.

The experiment was aimed at analysis of banana plantlets undergoing four levels of water stress, intended for detection and early prediction of drought conditions. 
The plantlets were grown in a greenhouse in northern Israel by Rahan-Meristem \cite{rahan2020website}. 
Each treatment category included $30$ plants, followed daily for $17$ consecutive days, and photographed once daily with a high resolution RGB camera and a low resolution thermal (infra-red) sensor.
All plants were cloned from the same original, and the experiment was conducted during September $2018$. 
All images were manually annotated with plant contours.
Image backgrounds were deliberately discarded, since they contain obstructive objects such as irrigation equipment, tagging labels, ground and cement. 

It should be noted that neither laymen nor experts can notice any difference of treatment in the images, and hardly from a sequence of images.
The single cue from the agricultural growers is that the appearance rate of the newest leaf in the plant is lower as the plant gets a decreased amount of water. 
In general, the RGB images display no visible difference, except appearance of new leaves once every $4$ to $6$ days.
The average temperature of the plants, extracted from the thermal images, does show a noticeable range variation, about $5.0^\circ$ between the extreme conditions.

Our main contributions include 
(1) creation of a RGB-T dataset in a controlled environment, planned specifically for analysis of plant structure and development, 
(2) proposing and comparing several methods for stress detection and prediction, 
(3) exploiting temporal image sequences to leverage the accuracy rates, and 
(4) employing shallow neural networks capable of feature extraction relevant to the issue at hand.
The following sections present related efforts in the field, description of the experiment and data acquisition, methods and strategies employed, results and conclusions.

\section{Related Efforts}
\textbf{RGB and thermal images.}
Several papers document the combined usage of RGB images, capturing light in the human visible range, and thermal images captured by infra-red (IR) or near infra-red (NIR) sensors.
Naturally, papers dealing with autonomous driving, motion tracking, and human pose estimation attract much attention in the computer vision community, hence the existence of RGB-T datasets in these fields is more common \cite{li2016learning,li2019rgb,shivakumar2019pst900,tang2019rgbt,palmero2016multi}.
In the agricultural domain, much research often rely on Unmanned Aerial Vehicles (UAV) and satellites collecting IR and NIR data for terrain mapping \cite{milella2017sensing,gago2015uavs,mulder2011use}, vegetation indexing of crop fields \cite{berni2009remote,genxu2012variability,bellvert2014mapping,raeva2019monitoring}, and canopy chlorophyll content \cite{delloye2018retrieval,gevaert2015generation,simic2018importance}.
Surveys related to UAV imagery all point to the importance of collecting and analyzing thermal and hyper spectral data for agricultural enterprises \cite{aasen2018quantitative,adao2017hyperspectral,abdullahi2015technology,barbedo2019review,kim2019unmanned,maes2019perspectives,mogili2018review}.
Simultaneous acquisition from multi-modal sensor systems is also indispensable for diagnosing plant stress at short-range \cite{chaerle2009multi,humplik2015automated}.

Thermal images are used in various agricultural applications - plant monitoring, yield estimation, crop maturity evaluation, irrigation scheduling, soil salinity detection, disease and pathogen detection,  and bruise detection \cite{ishimwe2014applications,khanal2017overview}.
Recently, several new RGB-T datasets were collected, specifically intended for applying computer vision methods to handle agricultural tasks, for example semantic segmentation of field patterns \cite{chiu2020agriculture} and estimation of weed growth stages \cite{etienne2019machine}.
Multi-modal data (LIDAR, radar, RGB, and thermal) was collected and studied for the purpose of obstacle avoidance in agricultural fields \cite{kragh2016multi,korthals2018multi}.
Other examples relying on thermal imagery include detection of tea disease \cite{yang2019tea}, detection of apple scab disease \cite{nouri2018near}, irrigation monitoring \cite{roopaei2017cloud}, and estimation of leaf water potential \cite{cohen2015crop}.

\noindent \textbf{Neural networks for plant phenotyping and stress prediction.}
Research in recent years has produced an abundance of practical implementations applying Convolutional Neural Networks (CNN) for many vision based tasks in the advancing field of precision farming and automated agriculture, usually fine-tuned for specific goals related to detection of crop traits and early stress prediction.  
CNNs are by far the most popular computational tool for feature extraction relevant to crop phenotypes. 
To list just a few CNN based examples, RGB-T is used for wheat ear detection \cite{grbovic2019wheat}, RGB and depth sensors for object detection intended for measurement of width and height of banana leaves, and height of banana trees \cite{vit2019length}, salt stress due to soil salinity using hyperspectral reflectance \cite{feng2020hyperspectral}, and leaf area, count, and genotype classification \cite{dobrescu2020doing}.
The importance of leaf segmentation for practically every plant phenotyping task is frequently mentioned \cite{scharr2016leaf,tsaftaris2016machine,ubbens2017deep}, leading to various tailored augmentation methods for accurate leaf segmentation, especially of Arabidopsis-thaliana \cite{kuznichov2019data,sapoukhina2019data,ward2018deep,zhu2018data}.  
Reflecting on current trends, phenotyping capabilities are already upgraded by research into the structure and optimization of neural networks. 
A discussion on reducing network size for flower segmentation is presented in \cite{atanbori2018towards}.
Lately, Dobrescu et al. presented a deep attempt to understand how CNNs reach decisions in leaf counting schemes \cite{dobrescu2019understanding}. 
Future studies will most likely continue to adapt neural networks, acquire larger amounts of multi-modal data, and aim at further automation of data collection and analysis.

\section{Experiment Outline and Data Acquisition}
\begin{figure}[htbp]
\centering
\includegraphics[height=8.5cm]{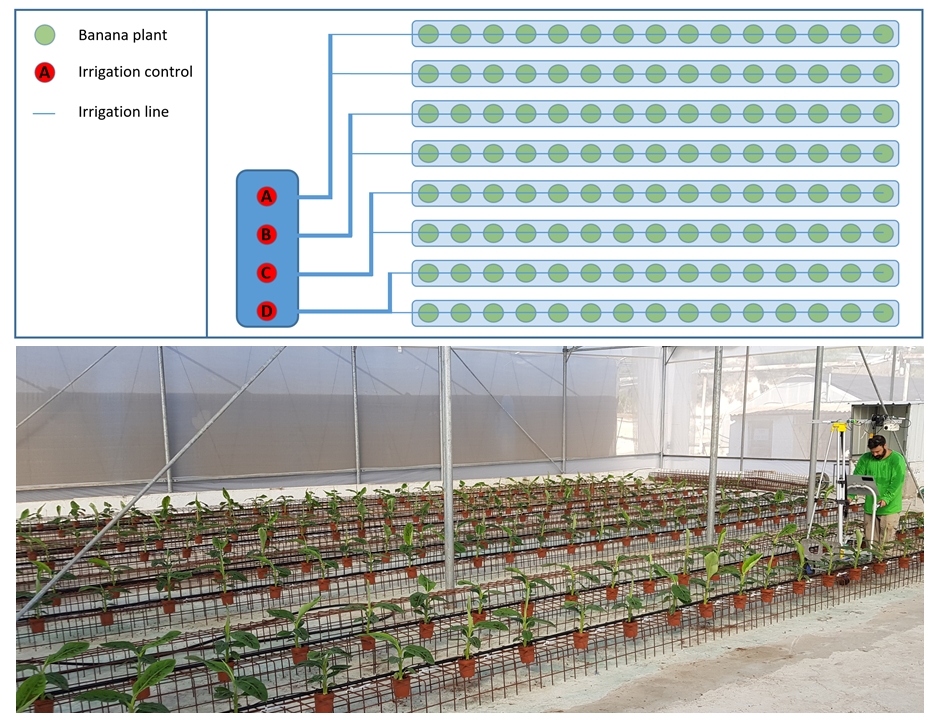}
\caption{Schematic representation of the experiment layout (top) and aerial view inside the greenhouse (bottom).}
\label{fig:experiment layout}
\end{figure}

\textbf{A stress detection venture.} 
The experiment was aimed at analysis of banana plantlets undergoing four levels of water treatment, and planned specifically for detection and prediction of drought conditions. 
One-month old plantlets were grown in a $1L$ pot in a commercial greenhouse.
The plantlets were positioned according to the scheme in Figure \ref{fig:experiment layout}. 
The four different treatments were labeled A,B,C,D for convenience, and each plant tagged with an ID, from $01$ to $30$.
During the first three days all plants received the same treatment; different water and fertilizer amounts were administered starting on the fourth day.
The plants in treatment A were watered and fertilized every day according to the normal commercial growing conditions ($100\%$), and plants undergoing treatments B, C and D were watered every day with lower amounts, $80\%$, $60\%$ and $40\%$, respectively, of the normal conditions. 
On the $17^{th}$ and last day of the experiment, irrigation of all plants returned to normal conditions, to support their recovery and verify that they are in a productive state for further usage.
Data collection time was between $07{:}00$-$10{:}00$, and watering of the plants at $13{:}00$.

\begin{figure}[htbp]
\centering
\includegraphics[height=3.2cm]{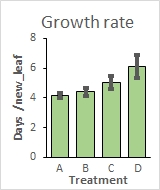}
\includegraphics[height=3.2cm]{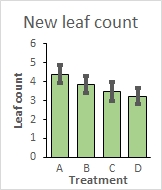}
\includegraphics[height=3.2cm]{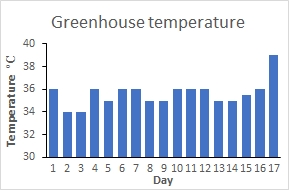}
\caption{Manual measurements during the experiment: Number of days between emergence of new leaves (left), count of new leaves during the entire experiment (middle), mean and standard deviation for each treatment, and measured temperature inside the greenhouse (right).}
\label{fig:manual measurements}
\end{figure}

\noindent \textbf{Data collection.} 
The RGB camera in use was a mobile Samsung SM-G930F, with pixel resolution $4032\times3024$, focal length $4mm$, and exposure time ${\sim}3.6ms$.
The IR sensor was a Opgal-ThermApp, pixel resolution $384\times288$, focal length $35mm$, and wavelength region $7.5 -14{\mu}m$ \cite{opgal2018camera}, outputting $16$ bits per pixel, each value an integer representing $100{\times^\circ}C$. 
All images are top views, taken approximately $1.2m$ from the top of plants. 
The temperature inside the greenhouse was measured once daily at a fixed position, being between $34$-$39{^\circ}C$. 
In addition, a log was kept, recording dates of leaf buddings in all plants. 
These manual measurements are displayed in Figure \ref{fig:manual measurements}.
Observing the leaf buddings turned out to be crucial; as will be explained later, our findings suggest a strong correlation between the rate of new leaf appearance and our stress prediction.

\section{Methods and Strategies}

\begin{figure}[htbp]
\centering
\includegraphics[height=2.2cm]{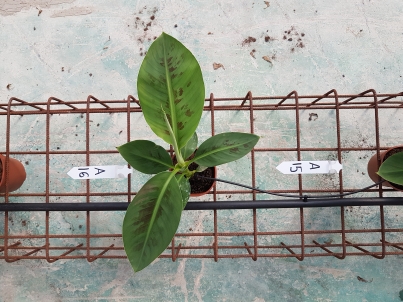}
\includegraphics[height=2.2cm]{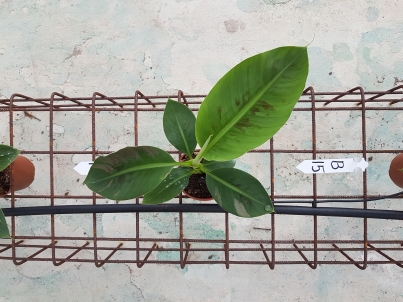}
\includegraphics[height=2.2cm]{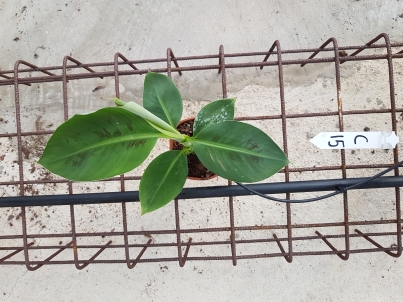}
\includegraphics[height=2.2cm]{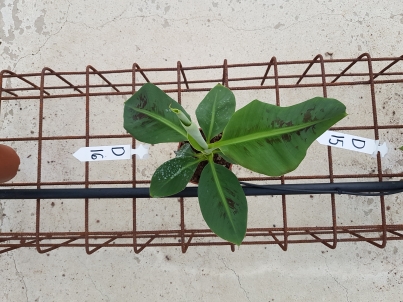}
\includegraphics[height=2.2cm]{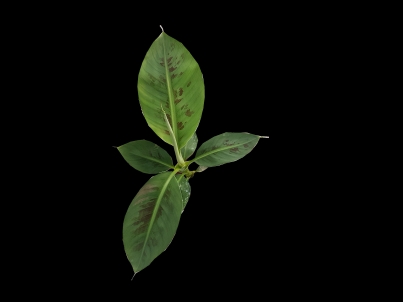}
\includegraphics[height=2.2cm]{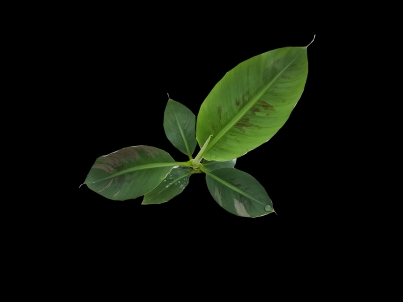}
\includegraphics[height=2.2cm]{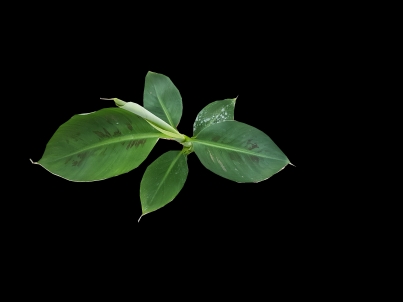}
\includegraphics[height=2.2cm]{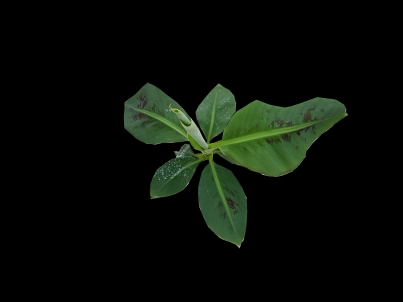}
\includegraphics[height=2.2cm]{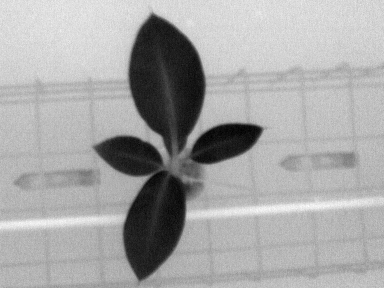}
\includegraphics[height=2.2cm]{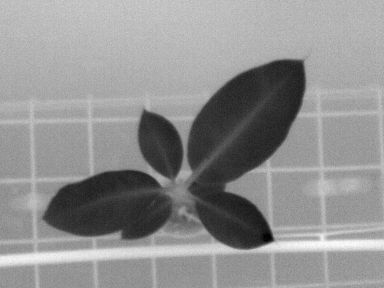}
\includegraphics[height=2.2cm]{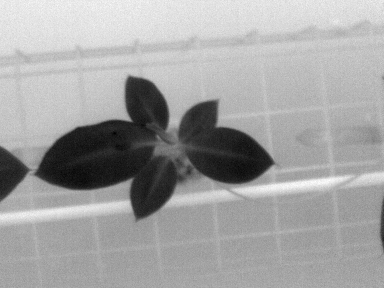}
\includegraphics[height=2.2cm]{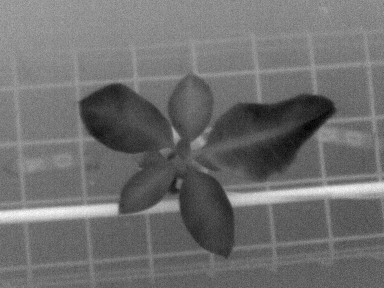}
\includegraphics[height=2.2cm]{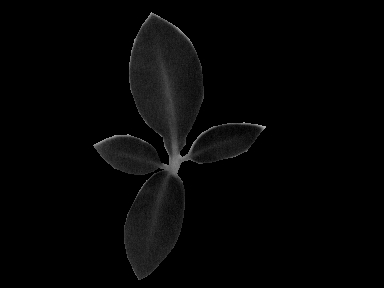}
\includegraphics[height=2.2cm]{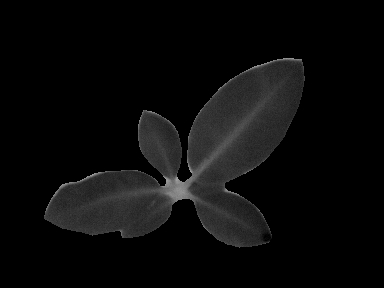}
\includegraphics[height=2.2cm]{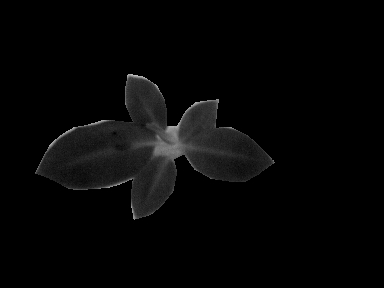}
\includegraphics[height=2.2cm]{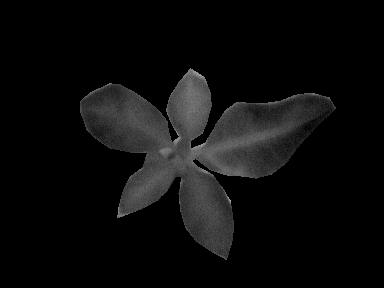}
\caption{Sample of original and segmented RGB images (top rows), original and segmented thermal images (bottom rows), from the eighth day of the experiment; A,B,C,D treatments (left to right), respectively. 
}
\label{fig:raw samples}
\end{figure}

\noindent \textbf{Pre-processing.}
All images were manually annotated with plant contours.
A custom built annotation tool was used, and contour points were marked on all images. 
Image backgrounds were deliberately discarded, since they contain obstructive objects such as irrigation equipment, iron support rails, tagging labels, ground and cement. 
Figure \ref{fig:raw samples} shows a sample  of raw input images, RGB and thermal, one from each treatment category, taken at the eighth day of the experiment.
Furthermore, the thermal images were passed through an erosion phase in order to discard any parts of the background that remained after the segmentation. 
A similar erosion in the RGB images was not deemed necessary, simply due to finer contour annotations, performed on the original high resolution images.

Thermal analysis commenced with measuring average temperatures on the plants, then comparing them with the average temperature of the plant contours. 
Several neural network architectures were then employed to use RGB and thermal images, separately, then combining both modalities.
The main boost to accuracy rates was obtained by inspecting consecutive images of the same plant from several days, three or more. 

Of the $30$ plants from each category, $26$ were used for training, the remaining used solely for testing. 
Plants with ID's $05,10,15,20$ served for test purposes, and in no way were seen by the training algorithms, neither RGB nor thermal. 
This separation of train/test was consistent throughout all the methods and prediction strategies.

\noindent \textbf{RGB data.}
Several network architectures were attempted, among them Resnet-50 \cite{he2016deep}, GoogLeNet \cite{szegedy2015going}, MobileNet \cite{howard2017mobilenets}, and VGG16 \cite{simonyan2014very}, or compacted versions of them, all widely used in classification problems. 
A Cifar-10 \cite{keras2020cifar10} based CNN architecture, trained from scratch, proved best. 
The deep pre-trained models resulted in performance far worse than training from scratch. 
This leads to the conjecture that general-purpose datasets are not well suited, in a transfer learning sense, to particular agricultural data.  
Our dataset is fairly small, so in order to reduce over-fitting, we used Keras built-in augmentation in our training process.
The augmentations we used include horizontal and vertical flips, width and height shifts, and rotations. 

\noindent \textbf{Thermal data.}
Similar to the RGB data, we first tried to fine-tune pre-built models like the GoogLeNet and Resnet-50 architectures, but here too, the results were far inferior to results from models we trained from scratch. 
We used a fairly similar model for this data compared to the RGB data model, but changed the order of convolutions because it proved to produce slightly better results.
Augmentations were applied during this learning process as well, horizontal and vertical flips, width and height shifts, and rotations.

\begin{figure}[htbp]
\centering
\includegraphics[height=6.5cm]{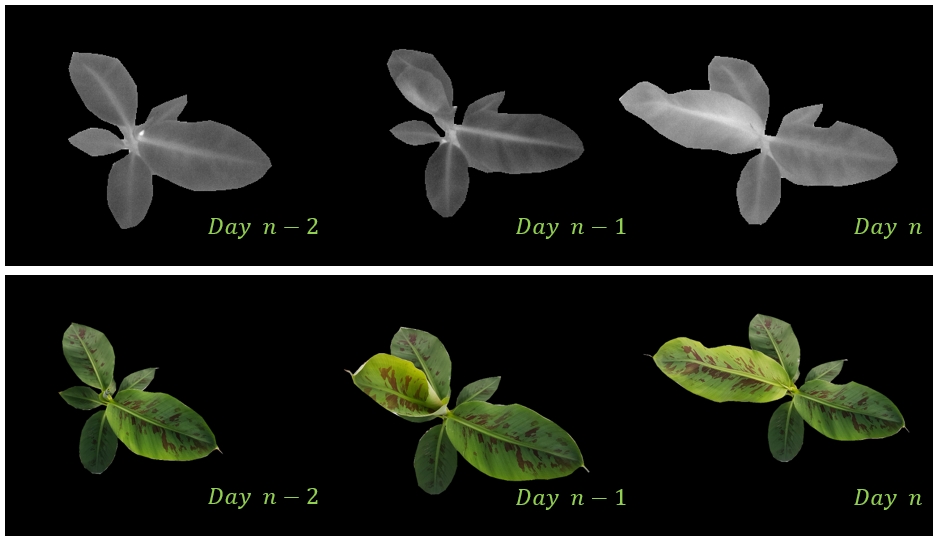}
\caption{Concatenation of images from $3$ consecutive days (treatment A, plant ID 14, days $13$-$15$). Notice the appearance of a new leaf bud at day {\it n-2} (a small cigar-looking leaf), its rapid growth between days {\it n-2} and {\it n-1}, and its final opening on day {\it n}.}.
\label{fig:triplet}
\end{figure}

Next, we tried training the same model with image sequences from $3$ consecutive days. 
Each sequence of $3$ images was concatenated and fed to our model as a single $288\times 1152$ pixel image to which we refer to as a {\em triplet}. 
We define \(triplet_n\) as constructed from images taken at days $n$-$2$, $n$-$1$, and $n$.
If $n$-$1$\(<\)$1$ or $n$-$2$\(<\)$1$ we simply duplicate the same image, so overall, we have the same amount of triplets as single images.
An example of such a triplet is shown in Figure \ref{fig:triplet}. 
According to our field observations, there is a strong correlation between the plant's drought stress and the rate in which new leaves appear. 
We believe that by using these triplets, our machine could deduce temporal features like leaf growth and size.

\noindent \textbf{Combination of RGB and thermal data.}
Because the thermal and RGB images are not aligned and have different resolutions, a model which trains on both types of images simultaneously may not perform well.
For that reason, a unique model was trained for each type of images separately.

\noindent \textbf{Prediction model.}
To classify a plant $x$, we pass $N$ consecutive thermal images (one for each day) through the thermal data model, and the corresponding RGB images through the RGB data model. 
For every image prediction, we save $P$ labels that match the thermal data model prediction, and $Q$ labels that match the RGB data model prediction, where \(\frac{P}{Q}\) is the ratio between the first model accuracy and the second model accuracy. For instance, if the thermal model's accuracy is $60$\% and the RGB model's accuracy is $90$\% then \(\frac{P}{Q} = \frac{60}{90} = \frac{2}{3} \rightarrow P = 2, Q = 3\).

To get the final prediction, we use a variation of the rolling average prediction method, meaning that our final prediction will be the highest probability (the one which appears the most times). 
In a formal manner, the prediction of plant $x$ is obtained as follows.
For each RGB image, we predict the label by passing it to the RGB model. 
We then duplicate the prediction $Q$ times,
\begin{eqnarray}
\mbox{pred}_{\mbox{rgb}}^* (RI_n) &=& (\mbox{pred}_{\mbox{rgb}} (RI_n),\cdots,\mbox{pred}_{\mbox{rgb}} (RI_n)) \in\mathbb{N}^Q, \;
\end{eqnarray}
where $RI_n$ is the RGB image of plant $x$ at day $n$. 
Similarly, each thermal image is fed to the thermal model to obtain the prediction. 
We then duplicate the prediction $P$ times,
\begin{eqnarray}
\mbox{pred}_{th}^* (TI_n)&=&(\mbox{pred}_{th} (TI_n),\cdots,\mbox{pred}_{th} (TI_n)) \in \mathbb{N}^P, \; 
\end{eqnarray}
where $TI_n$ is the thermal image of plant $x$ at day $n$. 
The combined prediction of this image at day $n$ is a concatenation of both label vectors
\begin{eqnarray}
\mbox{pred}^* (n)&=&\mbox{pred}_{th}^* (TI_n) \mathbin\Vert \mbox{pred}_{\mbox{rgb}}^* (RI_n) \in \mathbb{N}^{(P+Q)}, \; 
\end{eqnarray}
where $\mathbin\Vert$ is the concatenation operator. 
Then, we combine all prediction vectors from all $N$ images of the sequence to form a $1{\times}{N(P+Q)}$ buffer of labels,
\begin{eqnarray}
\mbox{buffer}&=&
 \mbox{pred}^* (1) \mathbin\Vert \mbox{pred}^* (2) \mathbin\Vert \cdots \mathbin\Vert \mbox{pred}^* (N) \in \mathbb{N}^{N(P+Q)}. \; 
\end{eqnarray}
For each possible label we sum the number of times it appears in the vector,
\begin{eqnarray}
C(l)&=&\sum_{l^\prime \in \mbox{buffer}} I\{l^\prime = l \}, \;
\end{eqnarray}
where $l^\prime$ is the predicted label and $l$ an element in the set of $4$ treatment labels.
We choose as final prediction the label with the most occurrences,
\begin{eqnarray}
\mbox{pred}(x)&=& \mbox{argmax}_{l\in \mbox{labels}} (C(l)).
\end{eqnarray}

We also tried to apply this prediction method on each model separately and save only one label per prediction in the buffer. 
This simple method enhanced the prediction accuracy on both models as can be seen in the results section.
Since the available data has a temporal quality, a Recurrent Neural Network (RNN) approach was also attempted but produced less satisfactory results.

\section{Results}
\begin{figure}[htbp]
\centering
\includegraphics[height=4.5cm]{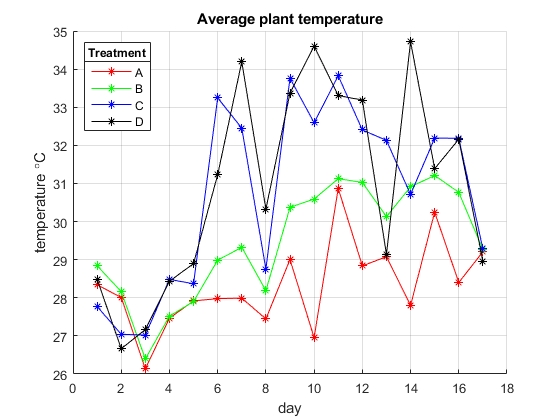}
\includegraphics[height=4.5cm]{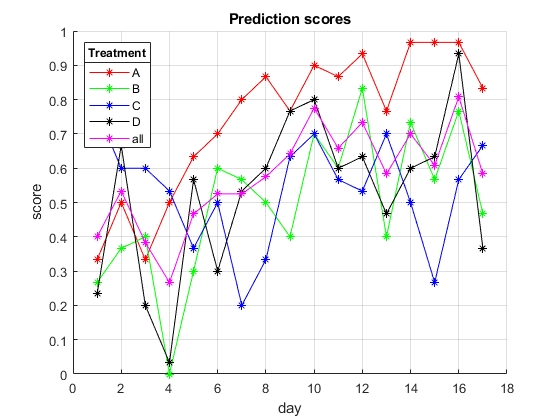}
\caption{Average daily plant temperatures for each treatment (left), and prediction accuracy rates obtained by plant and contour temperatures (right).}
\label{fig:basic results}
\end{figure}

\textbf{Preliminary analysis.}
Thermal analysis commenced with extraction of average temperatures on the plants compared with the average temperature of the plant contours. 
To this end, the contour temperature is defined as a narrow band of pixels between the original and slightly eroded image mask. 
Prediction accuracy by average plant temperature only, and difference between plant and contour temperature, resulted in ${\sim}50\%$ and ${\sim}60\%$, respectively, amongst the four categories.
In this preliminary analysis, a simple prediction scheme was employed: 
First, plant temperatures where obtained per treatment and per day from each thermal image, then averaged for each category.
Next, the predicted label of each image from the test set was defined as the treatment label (amongst A,B,C, and D) with average temperature closest to the test image. 
The prediction is considered true if the predicted label is the same as the ground truth label, and false otherwise.
Finally, average accuracies were computed for each day and each treatment category. 
Average plant temperatures and prediction accuracy at this stage are presented in Figure \ref{fig:basic results}.
Subtraction of the temperature measured once daily inside the greenhouse affected these results only slightly.
A note should be made that on the $17^{th}$ and last day of the experiment, irrigation of all plants was returned to normal conditions for all treatments, and the measured temperature on this day was significantly high (see Figure \ref{fig:manual measurements}), so these facts may explain the convergence of average temperatures on this day, as seen in Figure \ref{fig:basic results} (left).
The temperature drop of treatment D on the $13^{th}$ day may have been caused by a fault in the irrigation equipment.
This step already demonstrates that classification amongst the treatments can be achieved, to some degree, correlating the well known fact that plants receiving less water have a higher temperature. 
By using images from three consecutive days, the prediction accuracy was improved to ${\sim}70\%$.

\begin{figure}[htbp]
\centering
\includegraphics[height=4.0cm]{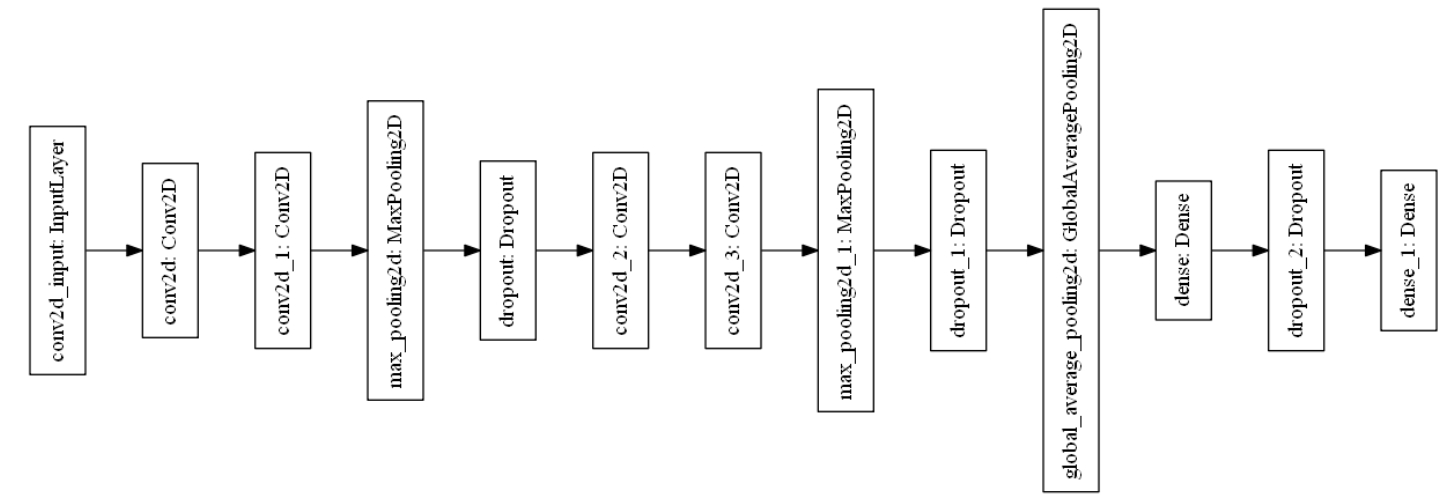}
\includegraphics[height=3.0cm]{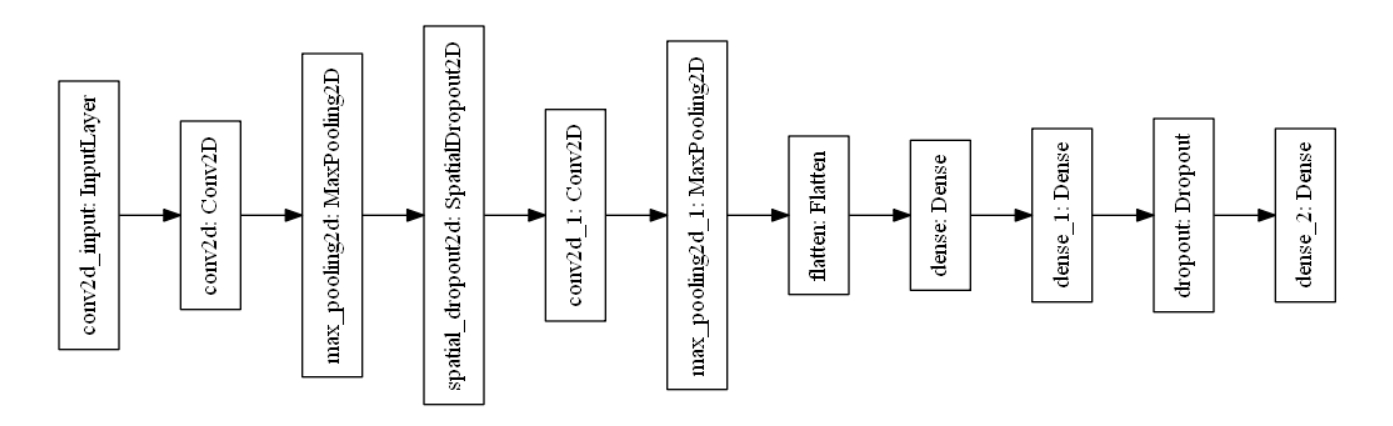}
\caption{The RGB (top) and thermal (bottom) architecture models; visualizations created by Keras visualising tool.}
\label{fig:network architectures}
\end{figure}

These results, moderately satisfactory, motivated us to apply deeper machine learning models, and integrate the temporal sequencing. 
The basic scheme is a network architecture composed of two convolution and max-pooling layers, followed by two fully connected layers, supplemented by $0.5$ dropout, with $32$ feature maps in the first convolution layer, and $64$ in the second.
The input is the original thermal images {\em as-is}, the RGB down-scaled to the same $384\times288$ resolution.
The RGB and thermal network architectures are presented in Figure \ref{fig:network architectures}. 
Again, implementation experience demonstrated that shallow networks trained from scratch performed better than deep pre-trained networks.

\noindent \textbf{Prediction by RGB images only.}
In this experiment we used a batch size of $24$ and trained for $100$ epochs using an ADAM optimizer.
As seen in table \ref{table:treatment prediction separate}, augmentation improved the results significantly, mainly because the dataset is rather small. 
Also, using a sequence of $3$ images from consecutive days achieved slightly better results compared to a single image.
One should not be surprised that the deep pre-trained models result in poor prediction, since they were explicitly trained to distinguish among semantic categories such as dogs, birds, cars, airplanes, etc., possibly different plant sub-species, but not amongst similar looking plant images. 
That said, the ${\sim}25\%$ prediction accuracy is even expected a-priory. 
In contrast, with limited amount of raw data, a shallow network accompanied with proper augmentation is expected to learn prominent features beneficial to the specific prediction task at hand, while deep training from scratch is an overkill.

\begin{table}[htbp]
\begin{center}
\caption{Treatment prediction accuracy using RGB and thermal images, separately; comparison of network architectures.}
\label{table:treatment prediction separate}
\begin{tabular}{ |p{4cm}||p{1.6cm}|p{1.6cm}|p{1.6cm}|p{1.6cm}|  }
\hline
 & GoogLeNet RGB & MobileNet RGB & Our model RGB & Our model Thermal\\ \hline
  Without augmentation & 25 & 36 & 72 & 40\\ \hline
  With augmentation & - & - & 82 & 62.5\\ \hline
  Triplets with augmentation & 25 & 32 & {\bf84} & {\bf72}\\ \hline
\end{tabular}
\end{center}
\end{table}

\noindent \textbf{Prediction by thermal images only.}
In this experiment we used a batch size of $32$ and trained for $1000$ epochs using a SGD optimizer. 
Here too, augmentation improved the results significantly and applying a sequence of $3$ images from consecutive days achieved better results than using a single image by a considerable margin.
However, results are far inferior to prediction by RGB images only. 
We believe there is more than one reason for that. 
Firstly, the thermal data is not very consistent; there were variations in the daily measured temperature inside the greenhouse, and the manufacturer specifies a ${\pm 3.0{^\circ}C}$ accuracy at $25{^\circ}C$ \cite{opgal2018camera}. 
These inconsistencies can be partially seen in Figure \ref{fig:basic results}. 
Secondly, the visual information gathered from a plant may have more discriminative information than its temperature. 
RGB contains $3$ channels, has much higher resolution, and also, today, low cost RGB cameras capture high quality info, while {\em quality-vs-cost} IR is still lagging behind.
For example it is easier to isolate each leaf in a RGB image than in its thermal image.
Due to these mentioned reasons, as well as less practical experience on learning from thermal images, the thermal network was trained for more epochs than the RGB network.
We found that $400$ epochs proved enough for converging on an asymptotic accuracy rate, and kept the process running in order to observe stable results before overfitting creeps in.

\begin{figure}[htbp]
\centering
\includegraphics[height=3.2cm]{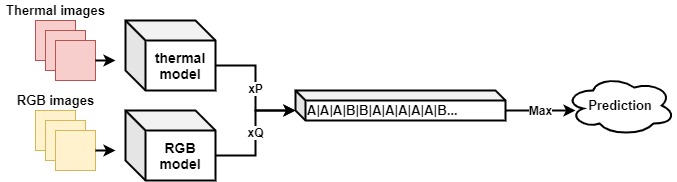}
\caption{The complete architecture. $x^P$ and $x^Q$ are $P$ and $Q$ prediction labels (thermal and RGB, respectively), possibly different for the same image, because of the dropout layer. These are concatenated into the label vector, which is {\em arg-maxed} to obtain the final prediction.}
\label{fig:thermal rgb architecture}
\end{figure}

\begin{figure}[htbp]
\centering
\includegraphics[height=5.0cm]{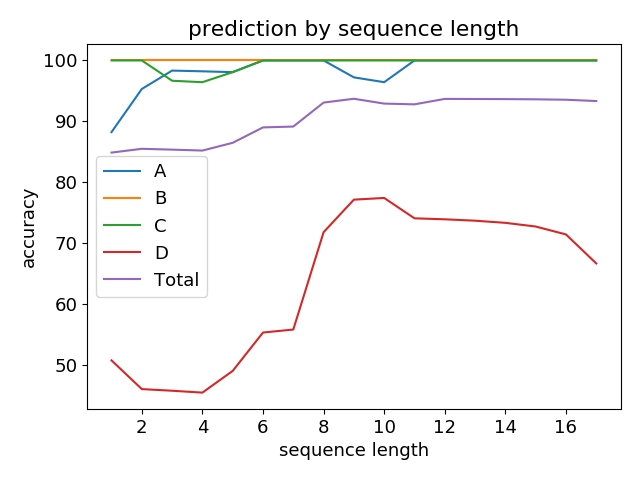}
\caption{Prediction accuracy by sequence length of the single RGB + single thermal framework.}
\label{fig:prediction by sequence length}
\end{figure}

\noindent \textbf{Integration of RGB and thermal data.}
In this experiment we integrated the models that performed best on their respective thermal and RGB data.
The entire architecture is depicted in Figure \ref{fig:thermal rgb architecture}.
We tried to combine the triplet models with the single images models and check which combination produces the best results. 
We also experimented with the sequence length $N$ to see how early it is possible to predict abiotic stress with a fairly good accuracy.
Table \ref{table:treatment prediction rolling average} shows the results of these experiments. 
Each row represents a single or combined model, and each column represents the number of consecutive images fed to the models. 
For example, column $8$ shows the mean prediction accuracy of the rolling average method over every sequence of $8$ consecutive images of all plants.
Figure \ref{fig:prediction by sequence length} is a graph visualization of prediction accuracy per treatment, obtained from one of the models, corresponding to the single RGB + single thermal line in table \ref{table:treatment prediction rolling average}.
Among all scenarios, best results were obtained by the single RGB + triplet thermal model. 

\begin{table}[htbp]
\tiny
\begin{center}
\caption{Accuracy of treatment prediction by rolling average, comparison of methods and sequence length.}
\label{table:treatment prediction rolling average}
\begin{tabular}{ |p{1.8cm}||p{5mm}|p{5mm}|p{5mm}|p{5mm}|p{5mm}|p{5mm}|p{5mm}|p{5mm}|p{5mm}|p{5mm}|p{5mm}|p{5mm}|p{5mm}|p{5mm}|p{5mm}|p{5mm}|p{5mm}|}
 \hline
 \multicolumn{18}{|c|}{RGB and thermal rolling average prediction accuracy} \\
 \hline
 Sequence length N& 1 & 2 & 3 & 4 & 5 & 6 & 7 & 8 & 9 & 10 & 11 & 12 & 13 & 14 & 15 & 16 & 17\\
 \hline
  Single RGB & 80.7 & 82.8 & 81.6 & 80.2 & 80.6 & 82.3 & 83.6 & 86 & 85.8 & 88.3 & 88.4 & 91.1 & 92 & 93.6 & 93.5 & 93.3 & 93.3\\
 \hline
 Single thermal & 69.4 & 71.1 & 67.7 & 65.7 & 68.6 & 72.5 & 78 & 81.8 & 82.6 & 83.8 & 87.3 & 83.5 & 87.3 & 85.1 & 87 & 86.6 & 80\\
 \hline
 RGB triplets & 78 & 78.2 & 80.2 & 80.6 & 81.6 & 82.8 & 83.6 & 84.6 & 85 & 86.5 & 87.3 & 87.3 & 87.3 & 87.2 & 87 & 86.6 & 86.6\\
 \hline
 Thermal triplets & 75.3 & 76.5 & 79.8 & 80.2 & 80.1 & 80.5 & 80.5 & 81.1 & 84.2 & 85.5 & 87.3 & 87.3 & 92 & 93.6 & 96.7 & 100 & 100\\
 \hline
 Single RGB + single thermal & 84.3 & 84.9 & 84.7 & 84.5 & 85.8 & 88.5 & 88.6 & 92.3 & 92.9 & 91.9 & 92.6 & 93.6 & 93.6 & 93.6 & 93.5 & 93.3 & 93.3\\
 \hline
 Triplet RGB + single thermal & 78.9 & 79.6 & 80.3 & 82 & 82.6 & 83.2 & 84 & 84.9 & 85.3 & 86.6 & 87.4 & 87.3 & 87.3 & 87.3 & 87.2 & 87 & 86.6\\
 \hline
 {\bf Single RGB + triplet thermal} & {\bf89} & {\bf89.5} & {\bf91.9} & {\bf91.8} & {\bf91} & {\bf91.4} & {\bf91.2} & {\bf92.3} & {\bf92.9} & {\bf94.6} & {\bf94.7} & {\bf100} & {\bf100} & {\bf100} & {\bf100} & {\bf100} & {\bf100}\\
 \hline
 Triplet RGB + triplet thermal & 78.8 & 80.3 & 82.5 & 84 & 84.8 & 85.7 & 86.1 & 87.4 & 87.4 & 87.4 & 87.3 & 87.3 & 87.3 & 87.2 & 87.1 & 86.6 & 86.6\\
\hline
\end{tabular}
\end{center}
\end{table}

\noindent \textbf{Discussion.} 
One may wonder about the seemingly {\em too-good-to-be-true} results presented in Table \ref{table:treatment prediction rolling average}, especially columns $12$-$17$ for the single RGB + triplet thermal architecture, and some explanation must be provided.
It is important to notice that the whole test set is fairly small, with only $16$ plants, $4$ from each of the $4$ categories (although the entire test set consists of $272$ thermal and $272$ RGB images, $4$ plants, $4$ treatments, $17$ days). 
Furthermore, as mentioned in the preliminary analysis, the na\"ive {\em average-per-treatment} prediction already reaches roughly $50\%$ accuracy, as opposed to random classification of $25\%$, amongst $4$ classes. 
Another thing worth mentioning is that the models were trained on all images from all days (except the test images). 
This means that the network can already be equipped to use this information for prediction while in some realistic situations we would not be able to train on {\em future} images.
As mentioned earlier, the {\em new-leaf-appearance-rate} is certainly highly correlated with plant stress. 
The triplets help capture this dependency after training the thermal and RGB models. 
In addition, the input to the complete architecture, which is a sequence of $N$ images from consecutive days, further increases its ability to detect high {\em new-leaf-appearance-rates}. 
In some cases, as seen in table \ref{table:treatment prediction rolling average}, the {\it single-image} scheme outperformed the {\it triplets}. 
We believe the reason for this is an inherent restriction on augmentations in the triplet arrangement; although a convolution based architecture is expected to be invariant to flips and rotations, the very arrangement of the triplets defies usage of horizontal flips, and rotations are also very limited, thus reducing generality of the training set. 

\begin{figure}[htbp]
\centering
\includegraphics[height=5.8cm]{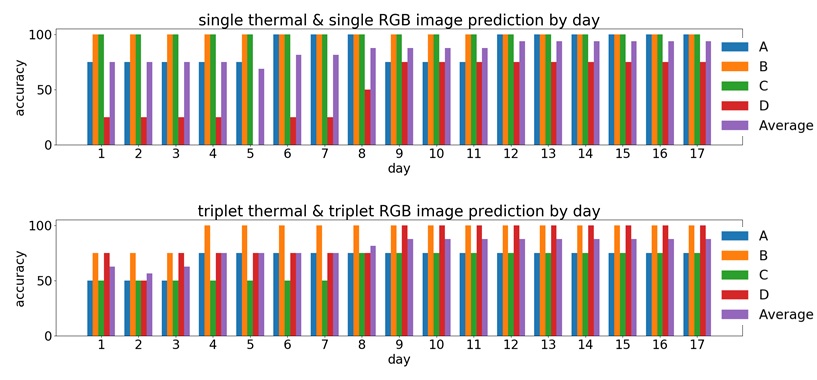}
\caption{Bar graph showing the prediction accuracy of each class as well as the average of all classes. 
Notice the total accuracy (in purple) rising as the day number goes up.}
\label{fig:prediction by day}
\end{figure}

Two other test scenarios were addressed, a binary normal {\em vs.} deficit stress, and capability of day-by-day stress prediction.
Table \ref{table:treatment prediction A vs all} presents an example of a binary dichotomy, normal treatment (A) {\em vs.} all others (B,C,D) suffering more or less from water deficit. 
No surprise here, the healthy plants are easily distinguished from the stressed.
In order to address the early prediction issue, we also supply the prediction accuracy per day, as opposed to any sequence of the last $N$ consecutive days.
Results of day-by-day prediction, obtained from combinations of single and triplet models, are presented in Figure \ref{fig:prediction by day}.

\begin{table}[htbp]
\tiny
\begin{center}
\caption{Binary prediction accuracy of normal condition (treatment A), {\em vs.} all other water deficit treatments.}
\label{table:treatment prediction A vs all}
\begin{tabular}{ |p{1.6cm}||p{4mm}|p{4mm}|p{4mm}|p{4mm}|p{4mm}|p{4mm}|p{4mm}|p{4mm}|p{4mm}|p{4mm}|p{4mm}|p{4mm}|p{4mm}|p{4mm}|p{4mm}|p{4mm}|p{4mm}|}
 \hline
 \multicolumn{18}{|c|}{RGB and thermal rolling average prediction accuracy} \\
 \hline
 Sequence length N& 1 & 2 & 3 & 4 & 5 & 6 & 7 & 8 & 9 & 10 & 11 & 12 & 13 & 14 & 15 & 16 & 17\\
 \hline
 Single RGB + Single thermal & 97.0 & 98.8 & 99.6 & 99.6 & 99.5 & 100 & 100 & 100 & 99.3 & 98.4 & 99.1 & 100 & 100 & 100 & 100 & 100 & 100\\
\hline
\end{tabular}\\
\end{center}
\end{table}

\section{Conclusion}
We have shown that machine learning methods and neural networks with comparatively small architecture have the capability to detect and predict stress conditions in plants.
Combination of RGB and thermal images has improved the results, and a temporal sequence of images has proven to obtain the best results.
Our results lead to the suggestion of combining multi-modal data, and acquisition of a controlled time-lapsed sequences of images.
We hope this modest contribution will serve the plant phenotyping community, by encouraging the ongoing efforts of carefully planned data collection, applying up-to-date computer vision methods, and analysis of stress detection and prediction.

\section*{Acknowledgment}
This research was partly supported by the Israel Innovation Authority, the Phenomics Consortium.

\par\vfill\par

\clearpage

%
%
\bibliographystyle{splncs04}
\bibliography{W40P14}

\end{document}